\documentclass[conference]{IEEEtran}

\IEEEoverridecommandlockouts
\usepackage{cite}
\usepackage{amsmath,amssymb,amsfonts}
\usepackage{algorithmic}
\usepackage{booktabs}
\usepackage{graphicx}
\usepackage{textcomp}
\usepackage{xcolor}
\def\BibTeX{{\rm B\kern-.05em{\sc i\kern-.025em b}\kern-.08em
    T\kern-.1667em\lower.7ex\hbox{E}\kern-.125emX}}
\begin{document}

\title{Efficacy of Large Language Models for Systematic Reviews}

\author{
    \IEEEauthorblockN{Aaditya Shah\IEEEauthorrefmark{1}\IEEEauthorrefmark{2}, Shridhar Mehendale\IEEEauthorrefmark{1}\IEEEauthorrefmark{2}, and  Siddha Kanthi\IEEEauthorrefmark{3}}
    \IEEEauthorblockA{\IEEEauthorrefmark{1}Illinois Mathematics and Science Academy, Aurora, Illinois \\
                      \IEEEauthorrefmark{3}Rick Reedy High School, Frisco, Texas}
    \thanks{\IEEEauthorrefmark{2} Both Shah and Mehendale contributed equally to this work; order of authorship is random.}
}

\maketitle

\begin{abstract}
This study investigates the effectiveness of Large Language Models (LLMs) in interpreting existing literature through a systematic review of the relationship between Environmental, Social, and Governance (ESG) factors and financial performance. The primary objective is to assess how LLMs can replicate a systematic review on a corpus of ESG-focused papers. We compiled and hand-coded a database of 88 relevant papers published from March 2020 to May 2024. Additionally, we used a set of 238 papers from a previous systematic review of ESG literature from January 2015 to February 2020. We evaluated two current state-of-the-art LLMs, Meta AI's Llama 3 8B and OpenAI's GPT-4o, on the accuracy of their interpretations relative to human-made classifications on both sets of papers. We then compared these results to a "Custom GPT" and a fine-tuned GPT-4o Mini model using the corpus of 238 papers as training data. The fine-tuned GPT-4o Mini model outperformed the base LLMs by 28.3\% on average in overall accuracy on prompt 1. At the same time, the "Custom GPT" showed a 3.0\% and 15.7\% improvement on average in overall accuracy on prompts 2 and 3, respectively. Our findings reveal promising results for investors and agencies to leverage LLMs to summarize complex evidence related to ESG investing, thereby enabling quicker decision-making and a more efficient market.
\end{abstract}

\begin{IEEEkeywords}
Large Language Models, Textual Analysis, ESG, Sustainability, CSR, Financial Performance
\end{IEEEkeywords}

\section{Introduction}

\subsection{Background and Motivation}

Environmental, Social, and Governance (ESG) investing has continued to grow, with nearly a 60\% increase in the use of ESG as investment criteria within the last two decades \cite{UMAR2020112}. Increasing awareness of sustainability issues has made ESG considerations a mainstream component for asset managers and institutional investors. The United Nations Principles for Responsible Investment now has over 609 asset owners holding USD \$121.3 trillion combined assets under management in 2021 \cite{https://doi.org/10.1111/joes.12599}. Additionally, the percentage of corporations releasing ESG reports has increased from 35\% in 2010 to 86\% in 2020 \cite{rouen_sachdeva_yoon_2024}.
 
However, there is no consensus on whether "ESG pays." Ľuboš Pástor, Robert F. Stambaugh, and Lucian A. Taylor \cite{PASTOR2022403}, for example, find that U.S. green stocks outperform brown stocks. In contrast, Juddoo et al. \cite{Juddoo2023} conclude impact investing strategies do not provide greater returns than traditional strategies. Many others have concluded that ESG does pay \cite{Huynh_Xia_2021, doi:10.1177/0149206320978451}, that ESG doesn’t pay \cite{Raghunandan2022, BOLTON2021517}, or that ESG has no impact on financial returns \cite{Laplume_Harrison_Zhang_Yu_Walker_2022, Amiraslani2023}.

In addition to this dispute in the industry, there are concerns regarding the volume, transparency, and usefulness of ESG reports released by corporations \cite{su151712731}. These inconsistencies complicate accurate measurement of a firm's ESG performance and its impact on financial outcomes. Large Language Models (LLMs) offer a promising solution in this context. LLMs can rapidly interpret and analyze vast amounts of text and numerical information, enabling them to improve the efficiency and accuracy of ESG data analysis and serve as a tool to help companies familiarize themselves with ESG standards \cite{su16020606}. Thus, LLMs offer the potential to reduce the complexity of the ESG investing landscape \cite{kumar2022}.

\subsection{Problem Statement}
With the rapidly changing nature of ESG investing and the large volume of available information, staying informed and rapidly summarizing new information is imperative. However, this requires extensive expertise and time, which not all industry players have access to. Therefore, it is essential to evaluate LLM tools in the context of ESG to determine if they can expedite the review process without compromising accuracy.

\subsection{Objectives}
The primary objective of this research is to evaluate the effectiveness of LLM tools (Llama 3 8B from Meta AI, GPT-4o from OpenAI, a "Custom GPT" chatbot from OpenAI, and a fine-tuned GPT-4o Mini Model from OpenAI) in conducting systematic reviews of the literature on the relationship between ESG performance and financial returns. By comparing the performance of LLMs to traditional manual systematic reviews, we evaluate the effectiveness of LLMs as a tool to help researchers and industry leaders efficiently interpret ESG information without compromising accuracy.

\subsection{Outcome and Significance}
We aim to contribute to two main strands of research: 1) the effectiveness of LLMs in summarizing and interpreting data and 2) the promise and pitfalls of systematic review. 

Our development of a "Custom GPT" and fine-tuned GPT-4o Mini model as well as the evaluation of two base model LLMs (Llama 3 and GPT-4o) highlighted the potential for LLMs to become a useful tool for interpreting ESG-related data. There were numerous occasions where all models excelled, such as with prompt 2 for papers classified as "accounting-based" or "both," with accuracies ranging from 87.1\% to 100\%. Additionally, the fine-tuned GPT-4o Mini model outperformed the base LLMs by 28.3\% on average in overall accuracy on prompt 1. The "Custom GPT" also outperformed the base LLMs but by 3.0\% and 15.7\% on average in overall accuracy on prompts 2 and 3, respectively. Ultimately, this signifies that LLMs can serve as a helpful resource in interpreting ESG literature, especially after fine-tuning with relevant data. 

Secondly, our study evaluated the use of LLMs in the context of systematic reviews to see if they can remove humans from the process. Systematic reviews are a key resource for advancing research, policy, and industry. However, the current best practices for systematic reviews are slow and labor-intensive, so much so that some reviews are no longer relevant by the time they are completed. Our study provides valuable insights into which models excel in specific contexts by assessing various LLMs in replicating systematic reviews. We highlight the potential for LLMs to transform systematic review best practices, given that the right prompt, model, and context have been selected.

\section{Previous Works}

In recent years, large language models (LLMs) have made significant progress in their Natural Language Processing (NLP) abilities, marking an important milestone in real-world applications of Artificial Intelligence (AI). These models can understand complex scenarios, analyze vast datasets, and generate relevant content, highlighting the immense potential within the sustainable financial industry \cite{chang2023surveyevaluationlargelanguage, wu2023bloomberggptlargelanguagemodel}.

In the context of finance, Araci \cite{Araci2019FinBERTFS} developed FinBERT, a pre-trained language model specifically fine-tuned for financial sentiment analysis. Leveraging BERT’s transformer architecture, FinBERT was trained on a large corpus of financial texts to grasp the nuances of financial language, enabling it to effectively classify sentiments in financial news and reports. FinBERT outperformed state-of-the-art models at the time, with an accuracy increase of 15

Gössi et al. \cite{gossifinbert} then fine-tuned the FinBERT model on a custom training set of 3,535 complex sentences from Federal Open Market Committee (FOMC) financial texts. This fine-tuned model saw significant accuracy gains over the previous FinBERT model, demonstrating the ability of LLMs to extract meaning from nuanced language and financial jargon.

Furthermore, Sarmah et al. \cite{Sarmah} investigated methods such as retrieval-augmented generation and the use of metadata to minimize hallucinations when extracting information from financial reports with LLMs. Meanwhile, Dolphin et al. \cite{dolphin2024extractingstructuredinsightsfinancial} employed chain-of-thought prompting to enhance the LLM’s sentiment analysis performance and rigorously validated output to ensure dependable results, increasing reliable information retrieval from a broader array of news sources by over 400\%.

In the specific context of this study—ESG analysis—AI models have been instrumental in redefining methods of scoring ESG performance for companies. Their effectiveness in interpreting financial data makes them powerful tools for analyzing vast amounts of unstructured data, such as company reports, financial statements, and news articles, to predict a company’s ESG performance \cite{zhao2024revolutionizingfinancellmsoverview}. Moreover, if AI models are trained on unbiased data, they can help eliminate human subjectivity and bias from ratings, reducing discrepancies across agencies \cite{zhao2024revolutionizingfinancellmsoverview}.

Del Vitto et al. \cite{DelVitto2023} evaluated white-box algorithms (Ridge and Lasso linear regression) and black-box algorithms (Random Forests and Artificial Neural Networks) on their ability to predict ESG scores using a training set of 13,052 firms stratified by industry sector and geographic region. Ridge and Lasso algorithms achieved root mean squared errors (RMSE) as low as 0.071, while artificial neural networks (ANNs) outperformed Ridge and Lasso within the Social pillar, achieving a lowest RMSE of 0.090 \cite{DelVitto2023}.

\section{Methods}

In this study, we assessed the performance of two base large language models (LLMs): Meta AI's Llama 3 8B\footnote{Accessed using the Groq API} and OpenAI’s GPT-4 Omni (GPT-4o). Our evaluation was conducted on two sets of papers: 88 papers published between March 2020 and May 2024 and a previously compiled set of 238 papers from January 2015 to February 2020 by Atz et al. \cite{Atz_VanHolt_Liu_Bruno_2021}. Each LLM was tested using nine specific prompts in a 3x3 experimental design (Fig. 5). Due to API token input limitations, Llama 3 8B was only provided with the abstracts of the papers, while GPT-4o was given the full PDF.

We also developed a custom version of ChatGPT, named "Custom GPT," using the "GPTs" feature on OpenAI's ChatGPT platform. This custom model was given comprehensive context plus chain-of-thought prompts for all three categories. Additionally, the 238-paper set with hand-coded labels was provided as reference data. The custom model, titled "ESG Returns Insights," was subsequently published on the ChatGPT platform. Additionally, we fine-tuned the GPT-4o Mini model on the 238-paper set and evaluated its performance on the new 88-paper set, benchmarking it against the standard GPT-4o Mini model. The fine-tuning process focused on Prompt 1, using the context prompt as the system instruction. 

\subsection{Data Collection}
We queried ProQuest, Science Direct, JSTOR, Google Scholar, and Social Science Research Network (SSRN) to find literature on ESG and financial performance published between March 2020 and May 2024. To ensure unbiased and high-quality papers were sourced, we filtered the results to include only papers from journals rated 3, 4, or 4* and classified as either ACCOUNT, FINANCE, ECON, ETHICS-CSR-MAN, or STRAT from the Chartered Association of Business 2021 Academic Journal Guide \cite{academic-journal-guide-2021}. We compiled a corpus of 98 papers and hand-coded each in a typical systematic review format. However, we removed 10 papers, all from the Journal of Business Ethics, as they were deemed not salient enough \cite{Atz_VanHolt_Liu_Bruno_2021}. Of the remaining 88 papers, 14 manuscripts are not peer-reviewed. In addition to the new papers, we downloaded an older set from January 2015 to February 2020 comprising 238 papers from Atz et al. \cite{Atz_VanHolt_Liu_Bruno_2021}.

\subsection{Data Classification on New Set}
The new data set labels followed the same structure as Atz et al.'s \cite{atz2021online} paper, as detailed in the online appendix \cite{Atz_VanHolt_Liu_Bruno_2021}. Three main questions were answered, and their responses were coded into the respective categories:

\begin{enumerate}
    \item Does the study conclude a relationship between sustainability and financial performance? (positive, negative, or mixed/neutral)
    \item How is financial success implemented? (market-based, \\accounting-based, both, or other)
    \item How is sustainability implemented? (ESG, E, S, G, CSR, or other)
\end{enumerate}

\noindent Two researchers independently coded the papers, achieving a rater agreement of 90\%, with a third researcher resolving the discrepancy.

\subsection{Prompt Categories and Definitions}
We split up three questions into three categories (1-3), which were structured incrementally with increasing context (A-C). The A-level prompts were basic and simply asked the questions and listed the possible classifications. The B-level prompts provided increased context from the Atz et al. \cite{atz2021online} codebook behind the meaning of the classifications, along with one or two examples. The C-level prompts included all of the information from the B-level and A-level but were designed to follow a chain-of-thought approach. They asked the model to take specific steps in determining its response. In addition, the model was asked to return a reasoning and confidence score for its prediction. This confidence score was derived within the prediction of the LLM and, therefore, should not be interpreted as a technical measure of confidence. These prompts were used for Llama 3 and base GPT-4o.

\subsection{Custom GPT Development and Model Fine-tuning}

The "Custom GPT" model was developed using the ChatGPT interface's "GPTs" tool, where we provided a set of instructions and relevant attachments for the default model context. The three chain-of-through prompts (1C, 2C, and 3C) were combined into one large set of instructions for the "Custom GPT." This way, following the input of a paper, all three prompts would be answered within a single response and without additional prompting. Each paper was tested in a new chat session to ensure that previous interactions did not influence the model’s responses. In addition to a specialized prompt, the model was provided the old corpus 238 papers as a labeled training sample.\footnote{The old corpus served primarily as a reference rather than modifying the model's parameters.}

Then, using the 238 papers in the old corpus of studies, we fine-tuned the smaller GPT-4o Mini model using the OpenAI API. We used a batch size of 1, 3 epochs, and a learning rate multiplier of 1.8. This model was only fine-tuned on Category 1 and instructed to return a prediction only for the answer to question 1.

\subsection{Data Processing and Analysis}
After obtaining the outputs from each LLM for the new and old datasets, we ensured that all results were standardized. We then compiled the results for further analysis. The primary focus was comparing the models' performance in correctly classifying the study findings and implementation methods across different prompts. We calculated the following metrics:
\begin{enumerate}
    \item Overall Accuracy 
    \item Sub-accuracy for Each Prompt Class
    \item F1 Score
\end{enumerate}

\noindent Given the disproportionate number of classes across prompts and the varying number of instances per class, we calculated the Weighted Power Mean (Equation 1) with \textit{p = 2} for the F1 scores for each predicted class. This approach provides a nuanced evaluation of our model’s performance, addressing the challenges of imbalanced data and varying class frequencies.

\vspace{10pt}
\begin{equation}
\text{Weighted Power Mean} = \left( \frac{\sum w_i x_i^2}{\sum w_i} \right)^{\frac{1}{2}}
\end{equation}

\vspace{5pt}

\section{Results}

\subsection{Old Data Set}

\begin{figure}
  \centering
  \includegraphics[width=\linewidth]{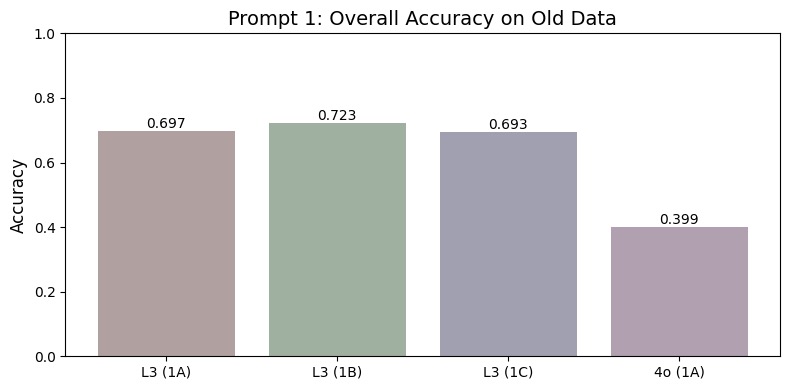}
  \label{fig:olddatafig}
  \caption{Accuracy of interpretation of Llama 3 on prompts 1A-C and GPT-4o on prompt 1A}
\end{figure}

\noindent We observed relatively no change in accuracy between the prompt variations for Llama 3 and noted a poor performance from base GPT-4o, nearly equivalent to guessing.

\subsection{New Data Set}

\begin{figure}
  \centering
  \includegraphics[width=\linewidth]{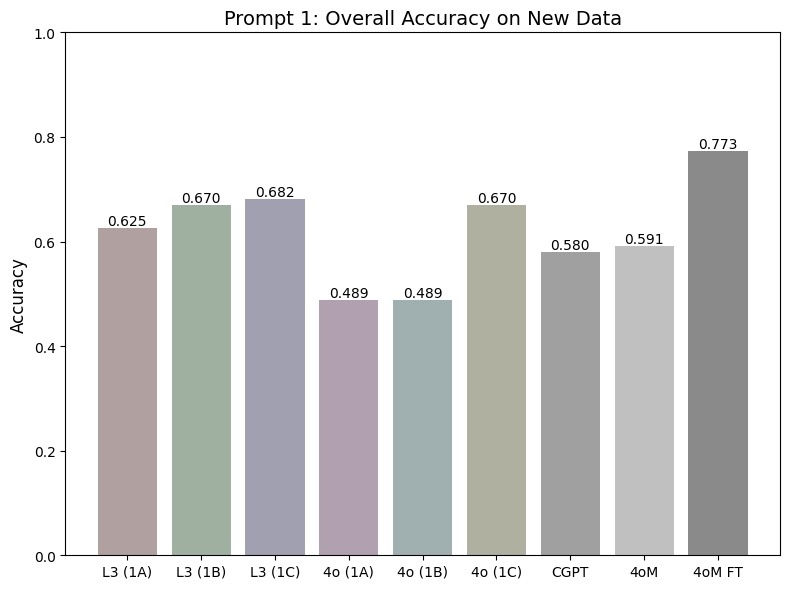}
  \label{fig:newdatafig}
  \caption{Accuracy of interpretation of Llama 3 and GPT-4o on prompts 1A-C.}
\end{figure}

\begin{table}
  \centering
  \caption{Accuracy of interpretation of Llama 3 and GPT-4o on prompts 1A-C overall and subsets}
  \label{tab:accuracy_1a_1c}
  \begin{tabular}{lcccc}
    \toprule
    Model & Overall & Positive & Negative & Mixed \\
    \midrule
    L3 (1A) & 0.625 & 0.695 & 0.565 & 0.438 \\
    L3 (1B) & 0.670 & 0.712 & 0.696 & 0.500 \\
    L3 (1C) & 0.682 & 0.729 & 0.696 & 0.500 \\
    4o (1A) & 0.489 & 0.373 & 0.870 & 0.250 \\
    4o (1B) & 0.489 & 0.390 & 0.870 & 0.250 \\
    4o (1C) & 0.670 & 0.644 & 0.826 & 0.500 \\
    CGPT & 0.580 & 0.576 & 0.696 & 0.375 \\
    4oM & 0.591 & 0.560 & 0.533 & 0.696 \\
    4oM FT & 0.773 & 0.780 & 0.733 & 0.783 \\
    \bottomrule
  \end{tabular}
\end{table}

The results show a substantial improvement in the GPT-4o base for prompt 1C, while Llama’s accuracy remained steady across all prompts. Both base LLMs showed poor performance on papers classified as "Mixed." GPT-4o performed its best in the negative category with a high accuracy of 87.0\%. The custom GPT-4o only outperformed GPT-4o in prompts 1A and 1B but not 1C and underperformed Llama 3 across all prompts. However, the custom GPT-4o Mini outperformed both base LLMs across all prompts. 

\begin{table}[h]
  \centering
  \caption{Accuracy of interpretation of Llama 3 and GPT-4o on prompts 2A-C overall and subsets}
  \label{tab:accuracy}
  \begin{tabular}{lccccc}
    \toprule
    Model & Overall & Market & Accounting & Both & Other \\
    \midrule
    L3 (2A) & 0.841 & 0.769 & 0.903 & 1.000 & 0.286 \\
    L3 (2B) & 0.852 & 0.385 & 1.000 & 1.000 & 0.286 \\
    L3 (2C) & 0.818 & 0.154 & 0.968 & 1.000 & 0.571 \\
    4o (2A) & 0.761 & 0.538 & 0.871 & 1.000 & 0.000 \\
    4o (2B) & 0.795 & 0.308 & 0.968 & 1.000 & 0.000 \\
    4o (2C) & 0.830 & 0.692 & 0.935 & 1.000 & 0.000 \\
    CGPT & 0.841 & 0.692 & 0.903 & 1.000 & 0.429 \\
    \bottomrule
  \end{tabular}
\end{table}

\noindent Base GPT-4o saw an increase in overall accuracy with the chain-of-thought prompt, whereas Llama 3 saw a decrease. Custom GPT-4o outperformed base GPT-4o across all three prompts, outperformed Llama 3 in 2C, and matched in 2B. All LLMs performed strongly on papers labeled as "Accounting-Based" or "Both" and poorly on papers labeled "Other." 


\begin{table}[h]
  \centering
  \caption{Accuracy of interpretation of Llama 3 and GPT-4o on prompts 3A-C overall and subsets}
  \label{tab:accuracy_3a_3c}
  \begin{tabular}{lcccccc}
    \toprule
    Model & Overall & ESG & E & S & G & CSR \\
    \midrule
    L3 (3A) & 0.773 & 0.828 & 1.000 & 0.200 & 0.933 & 0.786 \\
    L3 (3B) & 0.807 & 0.931 & 1.000 & 0.200 & 0.900 & 0.857 \\
    L3 (3C) & 0.818 & 0.931 & 1.000 & 0.400 & 0.900 & 0.786 \\
    4o (3A) & 0.602 & 0.414 & 0.333 & 0.200 & 0.800 & 1.000 \\
    4o (3B) & 0.727 & 0.621 & 0.667 & 0.400 & 0.900 & 0.929 \\
    4o (3C) & 0.693 & 0.793 & 0.000 & 0.200 & 0.800 & 0.857 \\
    CGPT & 0.852 & 0.966 & 0.333 & 0.500 & 0.967 & 0.786 \\
    \bottomrule
  \end{tabular}
  \begin{minipage}{\linewidth}
    \vspace{5mm}
    \footnotesize
    Note: The "Other" column has been removed from the table as it contained values of 0 for all models except for CGPT, which had a value of 0.5. Only two papers were within the "Other" category, accounting for 2.2\% of all papers.
  \end{minipage}
\end{table}

\noindent There was no notable difference when switching between prompts A-C for either Llama 3 or base GPT-4o. Custom GPT outperformed both base LLMs across all prompts. Llama 3 classified papers in the "ESG", "E", and "G" categories well. Base GPT-4o performed better with papers in the "G" and "CSR" categories. All three LLMs struggled with papers in the "S" and "Other" categories.

\begin{table}[h]
  \centering
  \caption{Confidence score given by Llama 3 and GPT-4o on chain-of-thought prompts}
  \label{tab:confidence_scores}
  \begin{tabular}{lcc}
    \toprule
    Prompt & Correct & Incorrect \\
    \midrule
    L3 (1C) & 0.866 & 0.832 \\
    L3 (2c) & 0.858 & 0.833 \\
    L3 (3C) & 0.880 & 0.872 \\
    4o (1C) & 0.937 & 0.924 \\
    4o (2C) & 0.969 & 0.924 \\
    4o (3C) & 0.966 & 0.926 \\
    \bottomrule
  \end{tabular}
\end{table}

\noindent There was no significant difference between the confidence scores when comparing both LLMs against each other and when comparing the scores for correct and incorrect classifications.

\begin{figure}[h]
  \centering
  \noindent\hrulefill
  
  \begin{flushleft}
  \noindent\textbf{Result:} Positive

  \noindent\textbf{Confidence:} 0.9

  \noindent\textbf{Reason:} The study concludes a significant and positive relationship between higher carbon emissions and higher stock returns across various sectors and countries.
  \end{flushleft}
  
  \noindent\hrulefill
  \caption{Base GPT-4o Misclassification Example on Prompt 1C}
  \label{fig:gpt4o_misclassification_example}
\end{figure}

\noindent GPT-4o overlooked the implication that higher carbon emissions equate to lower ESG, implying a negative correlation between ESG and financial performance. Misclassifications for similar reasons were prevalent across all three LLMs, occurring 9 times for base GPT-4o, 7 times for CustomGPT, and 5 times for Llama 3.\footnote{The fine-tuned GPT-4o Mini model only returned one-word classifications, therefore we were not able to determine the reason for misclassification.} 

\begin{table}[h]
  \centering
  \caption{A-C Sub-prompt Agreement on Prompts 1-3}
  \label{tab:prompt_agreement}
  \begin{tabular}{lcccc}
    \toprule
    Model & Overall & 1A-C & 2A-C & 3A-C \\
    \midrule
    L3 & 0.808 & 0.807 & 0.750 & 0.864 \\
    4o & 0.633 & 0.568 & 0.750 & 0.580 \\
    \bottomrule
  \end{tabular}
\end{table}

\noindent Both LLMs had a low agreement between predictions from prompts A-C, with Llama being consistent across all three prompts 80.8\% of the time, compared to 63.3\% for GPT-4o.

\begin{table}[h]
  \centering
  \caption{Weighted Power Mean of F1 score}
  \label{tab:weightedpowermean}
  \begin{tabular}{lccc}
    \toprule
    Model & 1 & 2 & 3\\
    \midrule
    L3 (A) & 0.639 & 0.852 & 0.796\\
    L3 (B) & 0.683 & 0.848 & 0.835\\
    L3 (C) & 0.689 & 0.811 & 0.844\\
    4o (A) & 0.488 & 0.784 & 0.632\\
    4o (B) & 0.492 & 0.797 & 0.749\\
    4o (C) & 0.679 & 0.842 & 0.757\\
    CGPT & 0.594 & 0.852 & 0.880 \\
    \bottomrule
  \end{tabular}
\end{table}

\noindent Table \ref{tab:weightedpowermean} shows the weighted power mean (with \( p = 2 \)) of the F1 scores for different LLM and prompt combinations across three scenarios.

\begin{figure}[h]
  \centering
  \includegraphics[width=\linewidth]{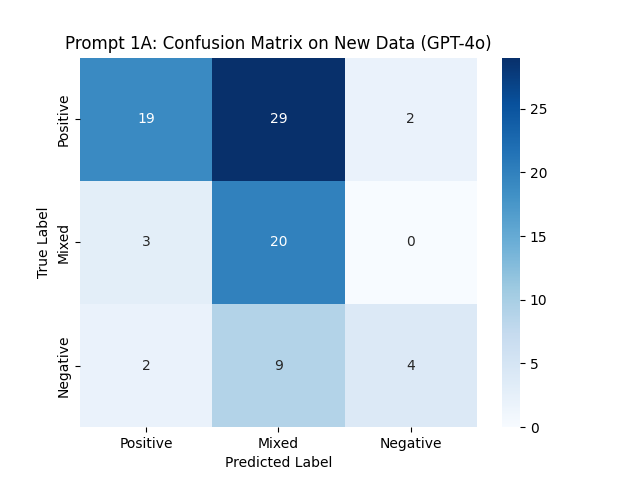}
  \label{fig:4o1a-cm}
  \caption{Confusion Matrix for GPT4o (1A)}
\end{figure}

Fig. 4 presents the confusion matrix for GPT-4o's performance on the new data set with prompt 1A. "Positive" samples were incorrectly classified as "Mixed" 58\% of the time. Overall, the model's responses were skewed to the "Mixed" category,  incorrectly classifying true "Positive" or "Negative" studies as "Mixed" frequently.

\section{Discussion}
\subsection{Base Llama 3}
Even though Llama only provided the abstracts of the papers, it outperformed GPT-4o on eight out of nine of the prompts. Furthermore, within Category 1, Llama 3 achieved the highest mean F1 score on Prompt C. Llama's overall accuracy ranged from 62.5\% to 85.2\%, with the model struggling on category 1, a phenomenon witnessed across all models. Additionally, Llama had an average prompt agreement of 80.8\%, making it reliant on needing a better prompt. 

Despite many papers in our corpus of 88 mentioning both metrics within the text, they typically used only one for their conclusions and thus were not classified as "Both." Llama 3 excelled in category 2 for "accounting-based" research papers, achieving accuracies between 90.3\% and 100\% and weighted power mean of F1 scores between 0.811 and 0.852, suggesting the model's ability to efficiently scan text for key details while filtering out irrelevant information.

Furthermore, Llama 3 classified papers coded as "ESG", "E", and "G" within category 3 well with accuracies ranging from 90.0\% to 100\% with one outlier of 82.8\% for prompt 3A under "ESG" classified papers. While many papers mention keywords such as "climate" and "environment," a broader analysis was necessary to comprehend the overarching theme and differentiate between "ESG" and the other individual categories. This highlights Llama 3’s remarkable ability to understand overall themes within the literature.

Overall, Llama 3’s performance within certain subcategories and consistent outperformance over base GPT-4o exemplify its viability as an LLM to assist during systematic review, but not entirely replace humans.
\subsection{Base GPT-4o}

Although base GPT-4o performed worse than Llama 3, it displayed highlights within certain prompts. For papers classified as "Negative" in category 1, the model displayed relatively strong accuracies ranging from 82.6\% to 87.0\%. Because papers classified as "Negative" often state this explicitly, the increase in accuracy may indicate that GPT-4o focuses more on searching for sentiment keywords throughout the paper. However, the model struggled with the "Positive" class and misclassified 58\% of "Positive" papers as "Mixed." Furthermore, similar to Llama’s increase in performance for category 2, GPT-4o performed better on the "Accounting-Based" paper, again highlighting the model’s ability to identify key information while ignoring the noise. Finally, GPT-4o was strong at classifying papers labeled as "CSR" within category 3, which further supports the idea that GPT-4o can scan for keywords and sentiment well because often when CSR was used as a metric by the paper, it was explicitly stated as "CSR" multiple times. The accuracy differences between GPT-4o and Llama 3 suggest a potential difference in training data between both LLMs. Additionally, the weighted power mean of the F1 score was the highest when using the chain-of-thought prompt, signifying that prompt design and structure play a significant role in the model outcome.

\subsection{Fine-Tuned Models}
The "Custom GPT" outperformed the base GPT-4o model for eight out of the nine prompts, only performing worse on prompt 1C. Compared to Llama 3, the "Custom GPT" outperformed on prompts 2C, 3A, 3B, and 3C, matched on prompts 2A, and underperformed on all category 1 prompts and 2C. Additionally, the Custom GPT achieved a higher mean F1 score than all other models in categories 2 and 3. 
Due to the "Custom GPT's" poor performance, in prompt 1, we fine-tuned a GPT-4o Mini model to excel within that category specifically. This new model outperformed all of the other models in category 1 and beat the standard GPT-4o Mini model by 30.8\%. This indicates that with fine-tuning and proper prompting/context, LLMs can effectively interpret ESG literature.

\subsection{Evaluation of LLMs as a Whole}

Ultimately, the application of LLMs within ESG shows potential. However, while LLMs can analyze vast amounts of data, the base model accuracy levels indicate that there is still room for improvement in terms of accuracy and deeper understanding. Specifically, the maximum accuracy achieved by base LLM models in determining the relationship between financial performance and ESG metrics was 68.2\%. When broken down into specific categories, models reached an accuracy of 77.6\% for financial metrics and 81.6\% for sustainability metrics. Notably, the base Llama 3 model consistently outperformed GPT-4o across these tasks, yet even Llama 3’s performance did not achieve an accuracy threshold that could replace human reviewers. This gap suggests that LLMs in their current form can support, but not fully automate, ESG analysis workflows.

To address these challenges, fine-tuning methods have been identified as a viable approach to enhance LLM accuracy and reliability. By training LLMs on ESG-specific datasets or tailoring them to a particular set of review standards, models can be refined to more accurately interpret and evaluate complex ESG data. With such fine-tuning, LLMs can potentially act as tools to expedite traditional systematic reviews, offering initial insights and analyses that streamline the review process while enabling human experts to focus on higher-level evaluation and decision-making. This approach positions LLMs as complementary assets to human reviewers rather than replacements, bridging the gap between automation and expert analysis in ESG-related applications.

\subsection{Future Works}
Our research underlines the importance of choosing the appropriate model architecture, prompting techniques, and contextual information to maximize the effectiveness of LLMs in ESG analysis. Notably, chain-of-thought prompting emerged as an effective method, enhancing accuracy by guiding the model through logical steps and clarifying its “thought process.” This approach provided greater transparency into the model’s reasoning, making chain-of-thought prompting a valuable area for further exploration and refinement in ESG applications.

Future research should prioritize the fine-tuning of LLMs using larger, more diverse, and ESG-focused datasets to improve their capability to interpret nuanced and complex information. Additionally, efforts should be directed toward developing frameworks that seamlessly integrate LLMs into existing systematic review workflows. With better-equipped models, researchers can employ LLMs to assist in essential review tasks, such as data extraction, synthesis, and reporting. These enhanced models have the potential to streamline the systematic review process while improving consistency and efficiency, ultimately contributing to the ongoing relevance and accuracy of ESG-related reviews and ensuring that findings remain aligned with industry standards over time.

\section{Abbreviations and Acronyms}\label{AA}

    \begin{itemize}
        \item L3: Meta AI's Llama 3 Model with 8B Parameters
        \item 4o: OpenAI's GPT4o Model
        \item 4oM: OpenAI's GPT4o Mini Model
        \item 4oM F: OpenAI's GPT4o Mini Model Finetuned on Corpus of Papers from 2015 to 2022 \cite{Atz_VanHolt_Liu_Bruno_2021}
        \item CGPT: Custom GPT using ChatGPT GPT-builder feature
        \item Mixed: Mixed/Neutral
        \item Accounting: Accounting-based metric of financial performance (ROA, ROE, etc.)
        \item Market: Market-based metric of financial performance (Tobin's Q, Jensen's Alpha, etc.)
        \item ESG: A holistic ESG score provided by third-party data providers
        \item E: Environmental factors exclusively
        \item G: Governance factors exclusively
        \item S: Social factors exclusively
        \item CSR: Corporate Social Responsibility 
        \item Prompt Category + Type (ex. 1C)
    \end{itemize}

\section*{Acknowledgment}
We thank Ulrich Atz for his guidance, review, and provision of OpenAI API computing credits. We also thank the seminar participants at the NYU Stern Center for Sustainable Business for their valuable feedback.






\bibliographystyle{ieeetr}
\bibliography{references}


\end{document}